\documentclass[journal,article,submit,pdftex,moreauthors]{Definitions/mdpi} 

\firstpage{1} 
\pubvolume{1}
\issuenum{1}
\articlenumber{0}
\pubyear{2023}
\copyrightyear{2023}
\datereceived{ } 
\daterevised{ } % Comment out if no revised date
\dateaccepted{ } 
\datepublished{ } 
\hreflink{https://doi.org/} % If needed use \linebreak
\pdfoutput=1 % Uncommented for upload to arXiv.org

\Title{A CNN-LSTM Architecture for Marine Vessel Track
Association Using Automatic Identification System
(AIS) Data.}

\TitleCitation{A CNN-LSTM Architecture for Marine Vessel Track
Association Using Automatic Identification System
(AIS) Data.}

\Author{Md Asif Bin Syed $^{1}$\orcidA{} , Imtiaz Ahmed $^{1}$*}

\AuthorNames{Md Asif Bin Syed, Imtiaz Ahmed}

\AuthorCitation{Syed, M.A.B; Ahmed, I.; }
\address[1]{%
$^{1}$ \quad Industrial and  Management Systems Engineering Department, West Virginia University}

\corres{Correspondence: imtiaz.ahmed@mail.wvu.edu;}

\usepackage{enumitem}
\newcommand*\rectangularbullet{\parbox[t]{6pt}{\mbox{\rule{6pt}{6pt}}}}
\usepackage{tcolorbox}
\usepackage{graphicx}
\usepackage{subcaption}
\usepackage{changepage}

\abstract{In marine surveillance, distinguishing between normal and anomalous vessel movement patterns is critical for identifying potential threats in a timely manner. Once detected, it is important to monitor and track these vessels until a necessary intervention occurs. To achieve this, track association algorithms are used, which take sequential observations comprising geological and motion parameters of the vessels and associate them with respective vessels. The spatial and temporal variations inherent in these sequential observations make the association task challenging for traditional multi-object tracking algorithms. Additionally, the presence of overlapping tracks and missing data can further complicate the trajectory tracking process. To address these challenges, in this study, we approach this tracking task as a multivariate time series problem and introduce a 1D CNN-LSTM architecture-based framework for track association. This special neural network architecture can capture the spatial patterns as well as the long-term temporal relations that exist among the sequential observations. During the training process, it learns and builds the trajectory for each of these underlying vessels. Once trained, the proposed framework takes the marine vessel's location and motion data collected through the Automatic Identification System (AIS) as input and returns the most likely vessel track as output in real-time. To evaluate the performance of our approach, we utilize an AIS dataset containing observations from 327 vessels traveling in a specific geographic region. We measure the performance of our proposed framework using standard performance metrics such as accuracy, precision, recall, and F1 score. When compared with other competitive neural network architectures our approach demonstrates a superior tracking performance.}


\begin{document}

\section{Introduction}

Tracking moving objects using both location and temporal information has enormous implications. It has the potential to improve maritime surveillance and security systems, as well as facilitate collision prevention and alert marine vessels to potential hazards. Tracking vessels is related to many active research domains such as track association \cite{9831001}, detection of anomalous vessels \cite{7363883,zhen_jin_hu_shao_nikitakos_2017}, trajectory prediction \cite{park2021ship,tang2022model}, clustering the moving objects with similar patterns \cite{li2018spatio,han2020dbscan}, and so on. Among them, track association, the concept of connecting unlabeled moving objects to their actual tracks, is comparatively less explored. The unavailability of reliable data, the absence of advanced computational techniques for exploiting big Automatic Identification System (AIS) datasets, the inherent complexity of the trajectory of the marine vessels, and the intrinsic behavior of these vessels can be cited as some of the underlying reasons for this setback. 

Spatio-temporal, geo-referenced datasets are rapidly expanding and will continue to do so in the near future due to technological advancements as well as social and commercial factors. The introduction of the AIS, which allows neighboring ships to communicate frequently with their location and navigation status via a radio signal, has enabled researchers to get their hands on datasets rich in spatio-temporal information \cite{United}. AIS data is collected from satellites and ground stations located all over the world. AIS data facilitates the mapping and characterization of maritime human and vessel activities, thus allowing for real-time geo-tracking and identification of vessels equipped with AIS. Hence, in addition to its initial application in collision avoidance, AIS is now a massive data source of unparalleled quality for diverse tracking tasks as well \cite{Marine}. 

AIS dataset contains the location and motion features of the vessels. Each data point or row in the AIS data file is represented by a time-sequenced node that contains the vessel's coordinates, speed, and traveling direction. Each node also has an associated time stamp indicating the data collection time. The nodes are arranged according to these time stamps. Additionally, each vessel in the AIS data is assigned a unique maritime mobile service identity (MMSI) number. This number helps to associate each node with the respective vessels. If this number is missing or purposefully concealed for a single node or stretch of nodes, a successful tracking algorithm should still be able to associate them to their true tracks. So far, several studies have used AIS datasets and developed algorithms to carry out trajectory prediction, anomaly detection, and clustering. However, most of these studies ignored some of the unique characteristics of these sea-going vessels. \cite{9831001,han2021modeling,zhang2020trajectory}. 

Vessel movement patterns can vary depending on diverse factors such as current location, shipping route, movement of other vessels, environmental factors, etc. So, no wonder, track association has been a challenging task for decades. Moreover, a ship can suddenly stop transmitting signals for various reasons ranging from unexpected equipment failure to intentionally hiding the true direction. Since a ship could change direction, increase speed significantly, or stop completely during this time, it would be significantly more difficult to correlate nodes to the correct lane after a long period of no communication. To overcome this challenge, it is important to mine the long-term temporal pattern of vessels.

Unlike land vessels, sea vessels do not change their positions and directions abruptly in the open sea and thus making it easier to track them. However, track association becomes more challenging when the vessels are close to a port. In a port, several vessels are docked close together, making it difficult to distinguish between them. Additionally, they frequently adjust their position in order to park or make room for other vessels. Consequently, it requires extra caution to track vessels near port \cite{9831001}. It is apparent that for successful tracking, it is important to take into account the vessel's location and previous movement patterns. In other words, the tracking algorithm should possess the ability to model both the spatial and temporal patterns hidden in the sequential observations.

A variety of techniques have been developed over the years for tracking moving objects. Multi-object tracking models \cite{blackman2004multiple,bar1995multitarget,reid1979algorithm} such as global nearest neighbor (GNN), joint probabilistic data association (JPDA), and multiple hypothesis tracking(MHT) are some of the most widely used approaches, with Kalman filtering at the core of these algorithms. However, Kalman filtering is more suitable for tracking linear motion and suffers in presence of non-linear trajectories with multiple overlapping tracks and missing data. These issues can be better handled by the physics-based models. While tracking they consider spatiotemporal patterns of the vessels and proved to be more effective for tracking maritime vessels than Kalman filtering \cite{9831001}. However, this framework uses only the most recent datapoint for track association thereby ignoring all the previous datapoints and the information hidden in them. These sequences often contain the long-term trajectory pattern of vessels and can help in future track association.

Deep learning architectures are proven to be quite effective in modeling nonlinear data and thus can be utilized for trajectory modeling as well. Recently, deep learning algorithms have been proposed to automatically learn hierarchical feature representations using raw spatio-temporal data quite similar to our case. Architectures such as convolutional neural networks (CNN) and recurrent neural networks (RNN) can capture spatial and temporal correlations present in the data. RNN is preferred to model any data that exhibit dynamic behavior, while CNN has proved to be an ideal approach for extracting spatial features from data \cite{wang2020deep}. The LSTM has been previously used in trajectory prediction \cite{park2021ship}. However, the presence of both spatial and temporal features in the vessel trajectories calls for an integration of both of these architectures. With the advancement of our research and computation capabilities, these joint architectures are evolving every year and more sophisticated frameworks are now taking place \cite{lu2021novel}. In light of these developments, we propose an integrated CNN-LSTM framework in this study. The academic contribution of our study is summarized as follows:

\begin{itemize}[label=\rectangularbullet]
  \item We develop an integrated 1D CNN-LSTM architecture for tracking marine vessels.
  \item We design and tune a CNN component architecture to extract the important spatial features from the dataset.
  \item We design and tune a LSTM component architecture to address the temporal behavior and long-term dependency among the nodes.
\end{itemize}

Our proposed approach has several strengths. First, it carries out a data pre-processing procedure to deal with issues such as incomplete dataset and gaps in data collection. Second, it utilizes a CNN architecture to help the dense LSTM layers in extracting important spatial features from the temporal data. Third, the data is fed into the LSTM architecture, a recurrent neural network that unravels and models the trajectory pattern of each specific vessel under consideration. Finally, the trained model is utilized to associate each incoming observation to its true track in real-time. We also compare the performance of our approach with other deep learning architectures using metrics such as F1 score, recall, and precision. 

The rest of the paper unfolds as follows. Section 2 highlights some of the existing research areas relevant to the track association problem based on the AIS or trajectory data. Section 3 discusses the data format, variables, and characteristics of the training and test data sets. It also describes the proposed methodological framework and network architecture for track association. Section 4 presents the performance metrics utilized and summarizes the comparative performance of our approach. Finally, we conclude the paper in Section 5.
%%%%%%%%%%%%%%%%%%%%%%%%%%%%%%%%%%%%%%%%%%
\section{Literature Review}
\label{sec:Literature Review}
%\subsection{Multiple Object Tracking (MOT) }

The process of monitoring an object over a sequence of frames to determine its location and direction is referred to as object tracking \cite{jimenez2022multi}. Both single and multi-object tracking techniques have been used in various applications, including computer vision problems, human behavior analysis, and security surveillance to name a few. Multi-object trackers (MOT) can be particularly useful in the context of marine security and surveillance, as they can combine information from multiple sensors like radar and sonar, instead of relying solely on AIS data and track several vessels in real-time \cite{9831001}.

However, applying MOT approaches to AIS data for track association is not straightforward and presents several challenging issues. Firstly, these approaches must appropriately handle the ambiguity surrounding the number of vessels. Sometimes this information is unknown and must be learned on the go. Secondly, object (vessel) appearance and disappearance are often uncertain, as they may emerge at unexpected times and locations and depart at the tracking boundary. Finally, multiple incidences of overlapping tracks and time gaps must be considered during the tracking process  \cite{9415072}. Over the past several decades, numerous research efforts have been undertaken to develop innovative and efficient solutions to multiple object-tracking problems. However, very few of them can handle all the issues posed by the AIS track association problem. The following subsections summarize these efforts into a few major areas namely sequential tracking models, physics-based models,  machine learning-based and hybrid models, and finally deep learning approaches.

\subsection{Sequential Tracking Models}

Sequential tracking algorithms, such as global nearest neighbor (GNN) \cite{blackman2004multiple} and joint probabilistic data association (JPDA) \cite{bar1995multitarget}, are commonly employed to update tracks based on the contribution of several objects. These algorithms utilize a cost assignment matrix to minimize costs and employ soft assignment, also known as track association probability, to achieve this goal. While Global Nearest Neighbor (GNN) \cite{blackman2004multiple} and Joint Probabilistic Data Association (JPDA) \cite{bar1995multitarget} focus on a single hypothesis for tracking objects, there are other techniques available. For example, multiple hypotheses tracking (MHT) \cite{reid1979algorithm} constructs a tree of hypotheses for each item and computes the likelihood of each track to determine the most probable combination of tracks. Random finite set (RFS) based approaches have also been utilized for tracking objects, as they are capable of handling the inherent uncertainty involved in the tracking process \cite{9415072}. The majority of Sequential tracking-based algorithms are based on the Kalman filtering (KF) approach \cite{capobianco2021deep} or its variations, which are frequently employed to track moving objects and provide information regarding their velocity and acceleration based on their position. However, the accuracy of KF is predicated on the assumption of linear motion, and it struggles to accommodate non-linear motion patterns. Furthermore, the KF framework has limited capacity for handling the distinct characteristics of vessel movements.

\subsection{Physics-based Models }

The physics-based approaches rely on mathematical equations to describe the motion of ships, taking into account factors such as mass, force, and inertia. These equations utilize physical laws to calculate the future motion characteristics of the ship \cite{best:1997,caveney:2007,semerdjiev:2000,khan:2005}. Such motion models can be useful for developing simulation systems to study ideal ship kinematic characteristics or even to train navigation systems. However, applying these models to track the trajectory patterns of multiple ships can be challenging. While these models can incorporate the spatiotemporal patterns of vessel movements \cite{9831001} in the learning process, they are still limited in nature, only considering the last known position to track vessels.

\subsection{Machine Learning and Hybrid Models }

These methods rely solely on historical data and employ machine learning techniques to learn from past information, enabling them to predict future positions when provided with a new feature vector. These prominent machine learning methods used in trajectory prediction studies include the Gaussian process, support vector machine, principal component analysis (PCA), etc. While these methods \cite{bartelmaos:2005, peng:2006, simsir:2009, joseph:2011,pallotta:2014} typically perform well in predicting immediate future positions, their prediction accuracy tends to decrease as the prediction time span increases. Furthermore, the performance of these models is highly dependent on the proper tuning of hyperparameters, which can be difficult to achieve. Additionally, they are not capable of processing long sequences and unraveling the spatial and temporal dependencies present in sequential observations. Hybrid approaches, on the other hand, combine physics-based models and machine learning models or different machine learning models to enhance the quality of the trajectory tracking process \cite{guo:2009,perera:2010,stateczny:2011,perera:2012,dalsnes:2018,tu:2020}. These approaches, however, are not free from the limitations imposed by the physics-based and machine learning models.

\subsection{Deep Learning-based Models}

Deep learning, which is a subclass of machine learning models, stands out from the rest due to its superior learning capabilities. In the context of marine vessel trajectories, neural networks have been widely used for their ability to process large datasets and discover long-term patterns hidden in vessel trajectories. These methods can identify all the possible trajectories a vessel can follow and reconstruct (predict) its trajectories for future time points \cite{volkova2021predicting}. Deep learning methods such as RNN \cite{capobianco2021deep},  (DBSCAN)-based long short-term memory (LSTM) model denoted as DLSTM \cite{yang2022deep}, and Variational Recurrent Autoencoder \cite{murray2021ais} were successfully utilized for trajectory tracking. To anticipate vessel motion and model its trajectory, neural networks are deployed as a nonlinear function for capturing vessel movement patterns. The design variables of these neural net-based frameworks are adjusted in real-time as the vessel travels \cite{borkowski2017ship,tang2022model,zheng2014trajectory}.
Instead of using the original features, these methods advocate the use of latent features derived from the neural network architecture. These latent features can capture the spatial patterns present in the data. Temporal ordering and attention maps are also proven to be effective for object tracking. \cite{xu2019deep}. 

Despite the decades of research on the object tracking problem, most existing tracking approaches rely on video datasets and conventional computer vision techniques. However, in our case, AIS dataset consists of numeric data with rich spatiotemporal information which demands an appropriate neural network architecture for handling this spatiotemporal aspect of the data. Furthermore, tracking multiple moving objects requires making several complex decisions on the fly, such as opening new tracks when a new object arrives, recognizing old tracks that have been inactive for a while, and closing existing tracks when they exit the tracking boundary. To address these challenges and fill the gap in the literature, in this work, we propose a 1D CNN-LSTM architecture for track association. This architecture is capable of processing multivariate sequential data and accurately assigning them to their correct tracks in real-time.

%%%%%%%%%%%%%%%%%%%%%%%%%%%%%%%%%%%%%%%%%%
%%%%%%%%%%%%%%%%%%%%%%%%%%%%%%% methodology 
%%%%%%%%%%%%%%%%%%%%%%%%%%%%%%%%%%%%%%%%%%
\section{Methodology}

This section presents our proposed methodological framework for the track association problem using a 1D CNN-LSTM approach. We discuss the framework's chronological steps in detail. The section begins by describing the AIS data, followed by an explanation of the data preprocessing techniques used to prepare it for our tracking algorithm. We then provide a summary of the traditional CNN and LSTM architectures, which are the component neural networks of our framework. Finally, we introduce the 1D CNN-LSTM architecture used in this study. The outline of the methodological framework is depicted below in Figure \ref{figure:1}:

\begin{figure}[h!]
\centering
\includegraphics[width=14 cm]{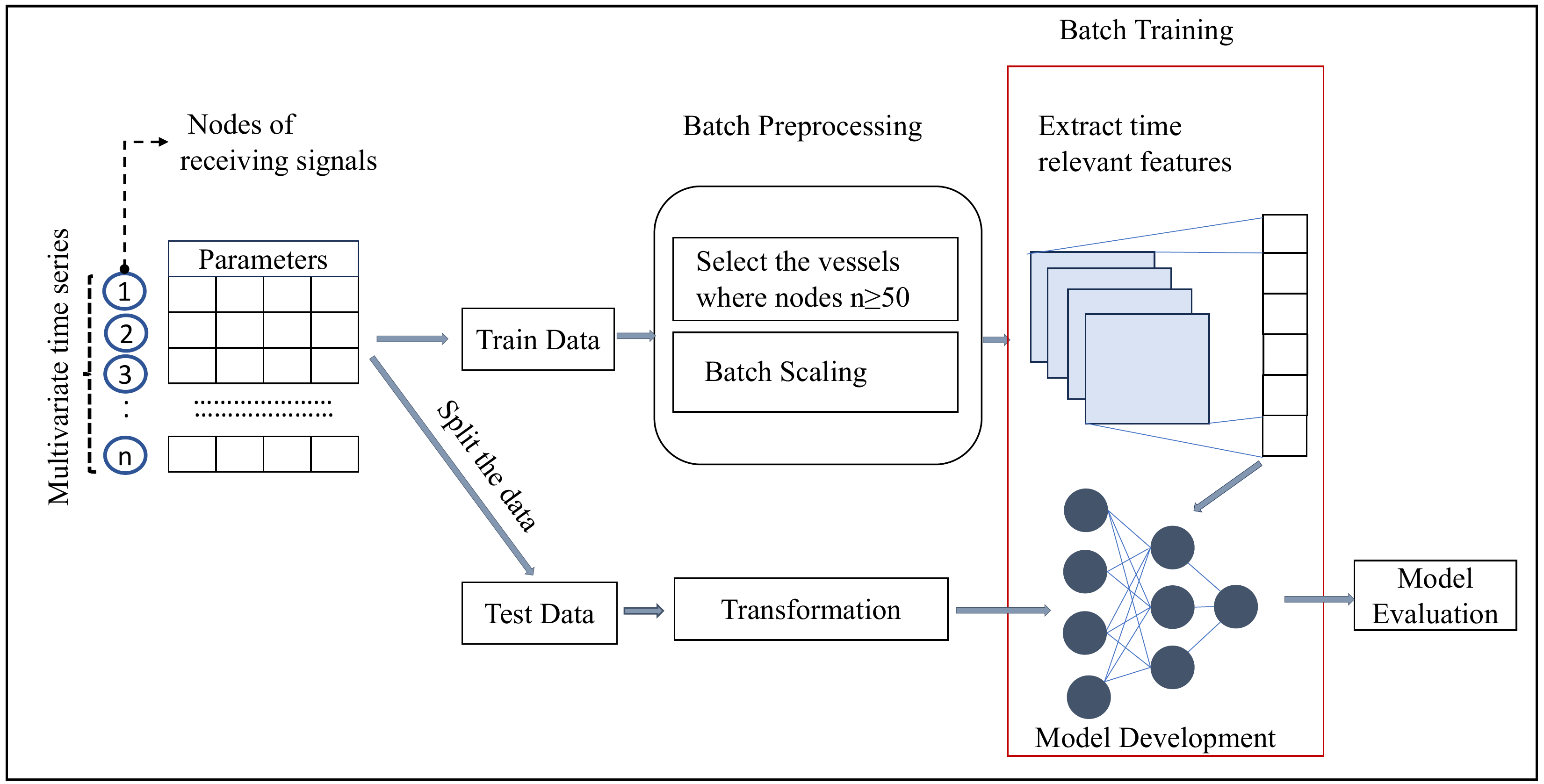}
\caption{Methodological framework for the track association problem.
\label{figure:1}}
\end{figure}  

\subsection{Data Description}
\label{subsec: data description}

This subsection introduces the Automatic Identification System (AIS) datasets which will be utilized to develop and test our proposed framework. The AIS dataset was procured from the National Automatic Identification System, maintained by the Coast Guard department of a specific country. When a signal is received in the AIS system, the incoming data is assigned a unique object ID and then verified by a monitoring officer to determine whether the vessel ID is already present in the existing database. The vessel ID, which is a combination of alphanumeric characters, serves as a unique identifier for each vessel in the system. If the received information matches a previously recorded VID, it is retained under the same identifier. However, if the vessel cannot be identified, it is assigned an identifier based on the timestamp. The AIS data is time-stamped and contains relevant information such as speed, the direction of movement, and location, which are expressed in terms of longitude and latitude. These timestamps serve as nodes for the track association problem.

The database utilized in this study was formatted as a CSV file, comprised of comma-separated values. To develop a track association algorithm, a subset of 327 vessels is chosen from the database, where the time intervals between signals are uneven. The irregular signal intervals present a challenge for traditional track association algorithms. To evaluate competing algorithms, approximately 25,000 data points are used for training, and 5,000 vessel IDs are reserved for testing. The VID information for these 5,000 test data points is intentionally omitted to evaluate the track association accuracy of the competing algorithms.

% Please add the following required packages to your document preamble:
% \usepackage{booktabs}
\begin{table}[]
\caption{AIS information in a CSV format dataset}\label{table:1}
\resizebox{\textwidth}{!}{%
\begin{tabular}{@{}lllllll@{}}
\toprule
\textbf{ID} & \textbf{VID} & \textbf{SEQUENCE\_DTTM} & \textbf{LAT} & \textbf{LON} & \textbf{SPEED} & \textbf{COURSE} \\ \midrule
1           & vessel 1     & 2020-02-29T22:00:01Z    & 37.8567167   & 23.53735     & 0              & 0               \\
2           & vessel 2     & 2020-02-29T22:00:01Z    & 37.9483      & 23.6410167   & 0              & 349.9           \\
3           & vessel 3     & 2020-02-29T22:00:01Z    & 37.9390233   & 23.6688483   & 0              & 228.3           \\
4           & vessel 4     & 2020-02-29T22:00:01Z    & 37.93884     & 23.6686333   & 0              & 0.1             \\
5           & vessel 5     & 2020-02-29T22:00:02Z    & 37.9314717   & 23.6804267   & 0              & 170.1           \\
6           & vessel 6     & 2020-02-29T22:00:02Z    & 37.9131117   & 23.5476617   & 0.1            & 33.3            \\ \bottomrule
\end{tabular}}
\end{table}

The dataset contains seven parameters, including a unique key for the database called object ID. The other parameters are vessel ID, timestamp (including date and time), latitude (expressed in degrees), longitude (expressed in degrees), speed (in tenths of knots), and course of direction (in tenths of degrees), as shown in Table \ref{table:1}. For clarity, Vessel IDs, which are lengthy combinations of alphanumeric characters unique to each vessel, have been replaced with labels such as vessel 1, vessel 2, and so on.

%\begin{figure}[t]
%\centering
%\includegraphics[width=12 cm]{images/trainingdata.png}
%\centering
%\caption{Visualizing vessels and associated tracks in the AIS dataset 
%\label{figrure:2}}
%\end{figure}  

%test 
\begin{figure}[t]
    \begin{minipage}{0.65\textwidth}
        \centering
        \includegraphics[width=\linewidth]{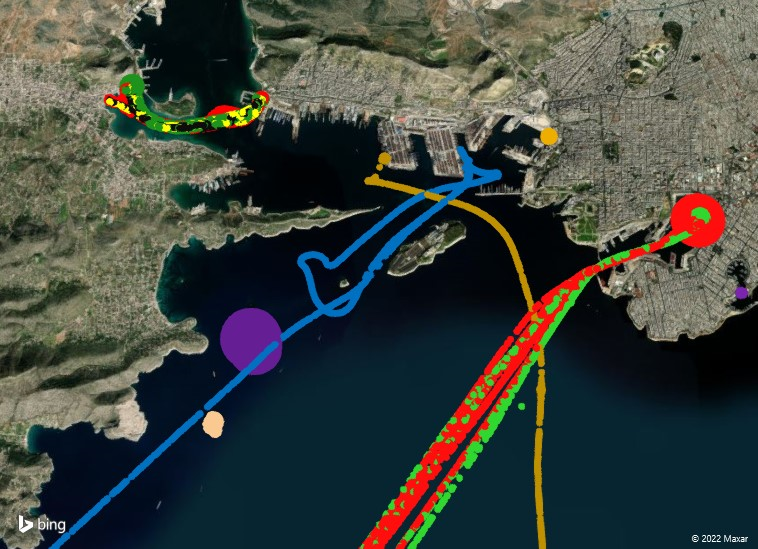}
    \end{minipage}
    \begin{minipage}{0.34\textwidth}
        \centering
        \caption{Visualization of vessels and associated tracks in the AIS dataset. Each color corresponds to a track from a different vessel. The map demonstrates the overlapping of the tracks of vessels which makes the track association problem challenging}
        \label{figrure:2}
    \end{minipage}
\end{figure}

For the model development, the timestamps, geographical information in the form of longitude and latitude, and dynamic parameters such as speed over ground and course of direction are selected as the primary input parameters. These selected parameters account for both temporal and spatial information. The vessel ID is designated as the target variable. In Figure \ref{figrure:2}, the intricacy of the track association problem we sought to investigate is visually demonstrated. The selected vessels exhibit overlapping tracks, creating a challenging scenario that mirrors the complexities present in real-world situations. Each color shown in the figure corresponds to a unique Vessel ID. In a subsequent section, the data is preprocessed to ensure that it is suitable for feeding into the proposed model.

\subsection{Filtering and Data Preprocessing}

\begin{figure}[h]
\centering
\includegraphics[width=0.7\linewidth]{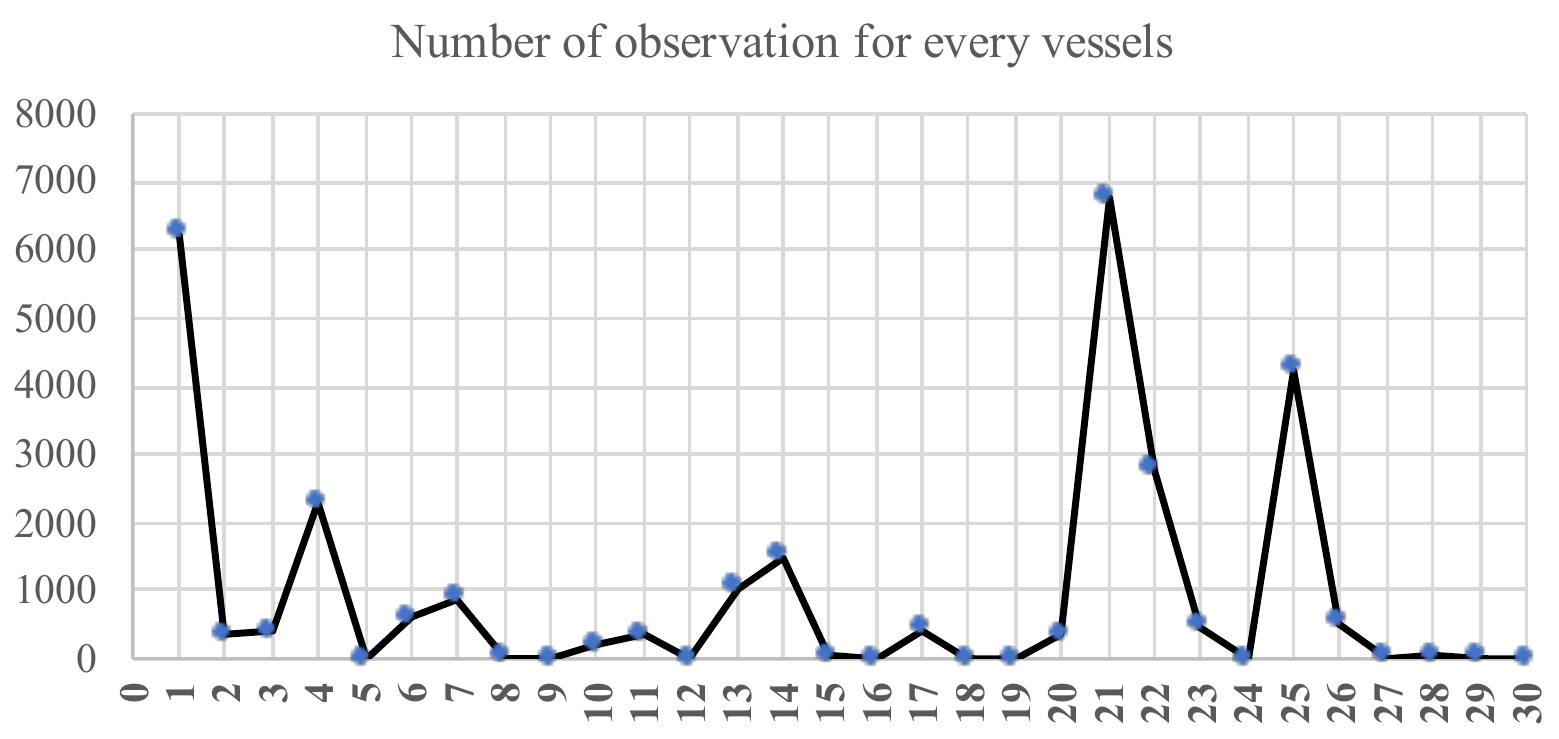}
\caption{Observation frequency for each of thirty vessels}
\label{fig:observations}
\end{figure}

The AIS data was collected and stored in a large CSV file, featuring data from 327 vessels. The preprocessing phase involves cleaning the data, where any timestamp containing a missing value will be removed. After cleaning, we observe that a significant number of vessels do not have enough data for training (as depicted in Table \ref{fig:observations}). To balance the bias-variance trade-off, we decide to limit the number of vessels used for training to thirty per batch. However, the vessel data distribution is also uneven across timestamps and it is challenging for the training process.
%%%test%%%%%%%

%%%test

To address this issue, we set a threshold of 50 observations to ensure that the model received sufficient data for training. Ultimately, data from 23 vessels and four variables are used, and they are normalized using the standard scaler equation provided below:

\begin{equation}\label{eqn:1}
z=(X-u) / s
\end{equation}

Here, $X$ represents the input data, $u$ is mean and $s$ represents the unit variance.

\subsection{Deep Learning Architecture }
\subsubsection{Convolutional Neural Network (CNN)}

Convolutional neural networks (CNNs) are a type of neural network that is designed to process and analyze data with spatial relationships. CNN has been widely utilized in various applications, such as audio signal pattern recognition, image processing, natural language processing, and time series prediction. The CNN architecture proposed by Lecun et al. \cite{lecun1998gradient} consists of an input layer, an output layer, and multiple hidden layers. These hidden layers contain convolutional layers, which perform dot products between the input matrix and the convolution kernel \cite{wikipedia}. 

We incorporate the CNN layer as the first layer of our hybrid deep learning architecture due to its ability to extract short-term patterns and dependencies among multiple input variables. They are particularly effective at extracting spatial patterns by using convolutional layers, which apply a set of learnable filters to the input data to detect features at different spatial locations. The convolution layer can be multi-dimensional, with its width t and height I representing the filter dimension. The output of the t-th filter, which operates on the input matrix X, is calculated below

\begin{equation}\label{eqn:2}
k_{\mathrm{t}}=\tanh \left(a_{\mathrm{t}} * X+b_{\mathrm{t}}\right)
\end{equation}

where $X$ is the input matrix, $b_{\mathrm{t}}$ is the bias matrix,  $a_{\mathrm{t}}$ represents the weight matrix and $k_{\mathrm{t}}$ is the output function. The output of the filter is then fed into the LSTM layer which is described in the following subsection.

\subsubsection{Long Short Term Memory (LSTM)}
Long short-term memory (LSTM) is a variant of Recurrent Neural Network (RNN) \cite{hochreiter1997long}. RNNs are a type of artificial neural network that can process sequential data by maintaining a memory of past inputs. In contrast to regular feedforward neural networks that process data in a single pass, RNNs are designed to handle data that has temporal dependencies, such as time series or sequences of text.  One of the challenges faced by conventional RNNs is the vanishing gradient problem, in which the gradient of the error function can either diminish or explode during backpropagation when it is propagated through multiple time steps.  

\begin{figure}[!htb]
\centering
\includegraphics[width=13.5 cm]{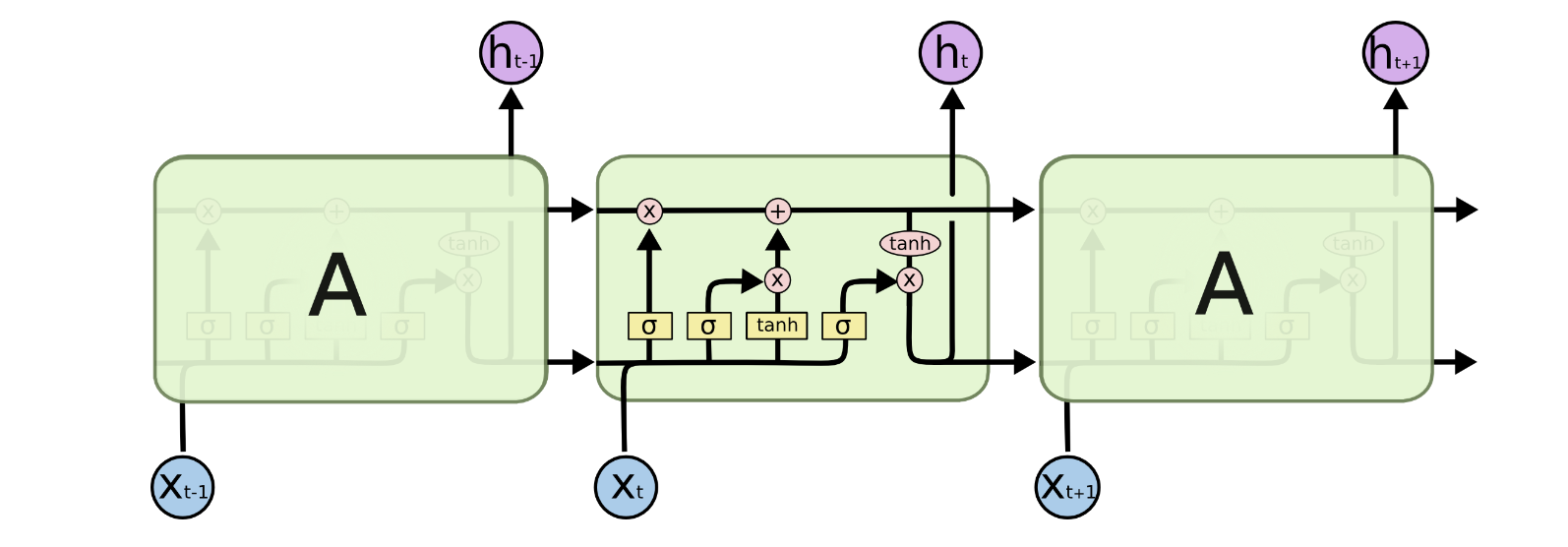}
%%\caption{Long Short Term Memory Architecture\label{fig2}}
\caption{Long Short Term Memory model schematic diagram}
\label{figure:4}
\end{figure}

However, the LSTM architecture overcomes this issue by incorporating a novel memory cell that regulates the information flow through the network. This mechanism selectively retains or discards information, which mitigates the vanishing or exploding gradient problem and allows learning long-term dependencies in sequential data. LSTM enables RNNs to perform long-term sequential prediction \cite{xie2020motion}. To enable sequential learning, LSTM employs three gates: a forget gate $f_t$, an input gate $i_t$, and an output gate $o_t$, which are depicted in Figure \ref{figure:4}. These gates control the flow of information through the memory cell. Elaborating on these three gates, the complex mechanism of LSTM architecture is explained below.

%lstm architecture

In LSTM, the forget gate plays a crucial role in determining which pieces of information to retain or discard from its existing memory, taking into account the arrival of fresh input. Let us denote the input time series as $X=\left(x_1, x_2, \ldots, x_t\right)$ and the hidden state of the memory cell as $H=\left(h_1, h_2, \ldots, h_{\mathrm{t}}\right)$. The forget gate takes the concatenation of previous hidden state $h_{t-1}$ and the current input $x_t$ as inputs, and produces a forget vector $f_t$ as output as in Equation \ref{eqn:3} \cite{blog}:

\begin{equation}\label{eqn:3}
\begin{aligned}
f_t = \sigma(W_f[h_{t-1}, x_t] + b_f)
\end{aligned}
\end{equation}

 Here, $\sigma$ denotes the sigmoid activation function, which is a nonlinear function that maps its input to a value between 0 and 1. The weight matrix $W_f$ represents the weights associated with the forget gate and determines how strongly the inputs affect the forget vector. $b_f$ represents the bias vector, which contains constant values that are added to the weighted sum of these inputs and can be thought of as the intercept term in the forget gate equation. The weights and the bias values are learned during the LSTM training process. 

LSTM employs the input gate to regulate the inflow of fresh data into the memory cell, which is composed of two components: the input activation gate and the candidate memory cell gate. The input activation gate dictates the extent to which the new input ought to be incorporated into the memory cell, whereas the candidate memory cell gate governs the portion of new data that must be retained in the memory cell.

By taking the previous hidden state $h_{t-1}$ and the current node $x_t$ as inputs, the input gate generates an input vector $i_t$ and a candidate memory cell vector $\tilde{c}_t$ in an LSTM. The operation of the input activation gate can be expressed using Equation (\ref{eqn:4}), which involves the weight matrix $W_i$ and bias vector $b_i$. Meanwhile, Equation (\ref{eqn:5}) illustrates how the candidate memory cell $\tilde{c}_t$ is formed by applying the hyperbolic tangent activation function ($tanh$) to the same set of inputs, using the weight matrix $W_c$ and bias vector $b_c$.

\begin{equation}\label{eqn:4}
\begin{aligned}
i_t & =\sigma\left(W_i\left[h_{t-1}, x_t\right]+b_i\right) 
\end{aligned}
\end{equation}
\vspace{-15pt}
\begin{equation}\label{eqn:5}
\begin{aligned}
\tilde{c}_t & =\tanh \left(W_c\left[h_{t-1}, x_t\right]+b_c\right) 
\end{aligned}
\end{equation}

The input vector and the candidate memory cell vector are then combined to update the previous memory cell $c_{t-1}$ (see Equation (\ref{eqn:6})). Here, $\odot$ denotes element-wise multiplication.

\begin{equation}\label{eqn:6}
\begin{aligned}
c_t & =f_t \odot c_{t-1}+i_t \odot \tilde{c}_t
\end{aligned}
\end{equation}

The final gate, the output gate, controls the flow of information from the current memory cell to the current hidden state, which is also the output of the LSTM at the current time step. First, the output vector $o_t$ is generated as in Equation (\ref{eqn:7}). 

\begin{equation}\label{eqn:7}
\begin{aligned}
o_t & =\sigma\left(W_o\left[h_{t-1}, x_t, c_t\right]+b_o\right) 
\end{aligned}
\end{equation}

Then the current hidden state, $h_t$ is obtained using Equations (\ref{eqn:10}) and (\ref{eqn:9}) where $\tilde{h}_t$ represents the candidate hidden state value for the current time step of the LSTM network. 

\begin{equation}
\begin{aligned}
\tilde{h}_t & =\tanh \left(c_t\right) 
\end{aligned}\label{eqn:10}
\end{equation}
\vspace{-15pt}
\begin{equation}
\begin{aligned}
h_t & =o_t \odot \tilde{h}_t
\end{aligned}\label{eqn:9}
\end{equation}

The following section provides an overview of the hybrid CNN-LSTM model architecture, including a description of the trainable parameters utilized in each layer and the hyperparameters chosen for the model.

\subsubsection{1D CNN-LSTM Architecture} 
% this is for subfigure images 

\begin{figure}[!htb]
\centering
\includegraphics[width=14cm]{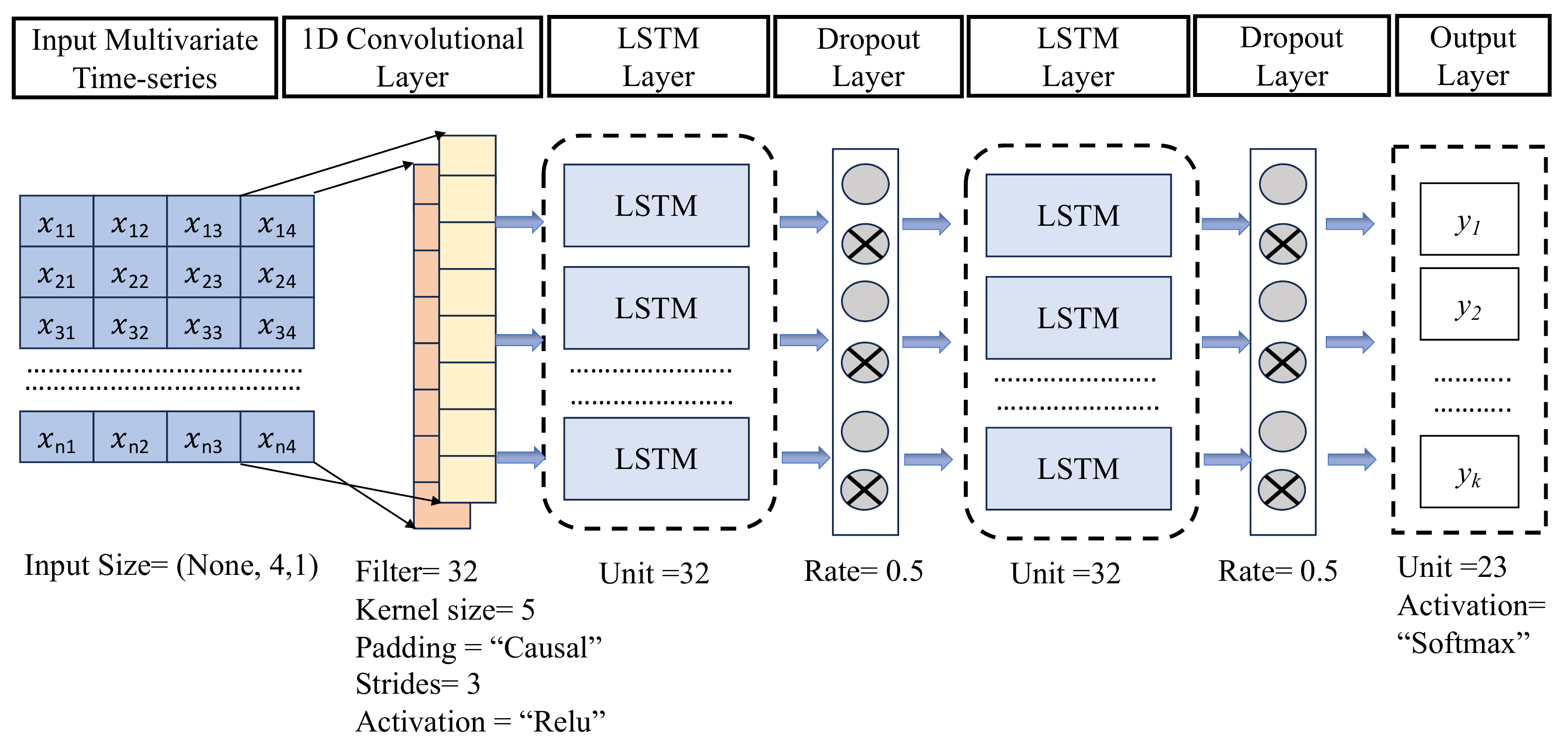}
%%\caption{Long Short Term Memory Architecture\label{fig2}}
\caption{The proposed 1D CNN-LSTM architecture}
\label{figure:5}
\end{figure}

To address the track association problem, it is essential to devise a model that can effectively incorporate both temporal and spatial information. The CNN-LSTM model, as illustrated in Figure \ref{figure:5}, represents a comprehensive solution that integrates one CNN layer and two LSTM layers to capture the underlying temporal and spatial dependencies present in AIS data. 

The proposed hybrid architecture takes an input vector represented by 
\[
X = 
\begin{bmatrix}
x_{11} & x_{12} & x_{13} & x_{14} \\
x_{21} & x_{22} & x_{23} & x_{24} \\
\vdots & \vdots & \vdots & \vdots \\
x_{n1} & x_{n2} & x_{n3} & x_{n4}
\end{bmatrix}
\]

with dimensions (None, 4, 1) and passes it through a convolutional layer. The first dimension, denoted as ``None," signifies the dynamic batch size commonly used in Keras implementation. Keras is a Python-based open-source neural network library that simplifies the prototyping of neural networks. The input vector consists of four variables, expressed as one-dimensional arrays, necessitating the use of a 1-dimensional Convolutional Neural Network (1D CNN) in this architecture. The 1D CNN operates with a kernel that can traverse only in a single direction. The convolutional layer processes the input and generates an output of size (None, 2, 32) (see Table \ref{tab6}). Here, 32 represents the filter size, while the layer utilizes a kernel size of 5, strides of 3, ``Relu" activation function, and ``casual" padding. The output volume's depth determines the number of neurons in a layer connected to the same region of the input volume, with each neuron trained to activate various input features. The stride allocates depth columns around the width and height of the output. The output of the convolutional layer is then passed to an LSTM layer that employs an input, output and forget gate mechanism. As discussed earlier, the LSTM layer produces an output and a hidden state value.

%% input output table
\begin{table}[ht!]
\caption {Parameter setting for the hybrid 1D CNN LSTM architecture}\label{tab:2}
\centering
\begin{tabular}{|l|l|l|}
\hline \textbf{Layer name} & \textbf{Output shape} & \textbf{Number of trainable parameters} \\
\hline Input layer & (None, 4, 1) & Not applicable \\
\hline Convolutional layer 1 & (None, 2, 32) & 192 \\
\hline LSTM layer 1 & (None, 2, 32) & 8320 \\
\hline Dropout & (None, 2, 32) & 0 \\
\hline LSTM layer & (None, 2, 32) & 8320 \\
\hline Dropout & (None, 2, 32) & 0 \\
\hline Dense layer  & (None, 23) & 759 \\
\hline
\end{tabular}
\label{tab6}
\vspace{.7 cm}
\end{table}

To avoid overfitting, a Dropout layer has been added between two LSTM layers. The Dropout technique randomly eliminates certain nodes at a specified probability during each weight update cycle. In this case, a dropout rate of 50\% is used, given the large amount of data being trained. The second LSTM layer generates an output with dimensions (None, 2, 32). The final layer in the architecture is a Dense layer that comprises an array of neurons. Each neuron receives inputs from all the neurons in the preceding layer. The output layer utilizes the Softmax activation function to handle categorical variables. 23 neurons are selected for the output layer, corresponding to the 23 vessels in each batch. Table \ref{tab:2} lists all the layers and trainable parameters.
%testing and training

To demonstrate the efficacy of the proposed hybrid CNN-LSTM method, extensive experiments are carried out using four distinct methods, including our own approach. The hyperparameters used for training the model are listed below:

\begin{table}[ht!]

\caption {Hyperparameter tuning for the hybrid 1D CNN LSTM architecture}
\centering
\label{tab:3}
\begin{tabular}{cc}
\hline Parameters & Value \\
\hline Convolutional layer filters & 32 \\
Convolutional kernel size & 5 \\
Convolutional kernel stride & 3 \\
Convolutional layer activation function & ReLU \\
Convolutional layer padding & Causal \\
Number of LSM hidden cells & 32 \\
Number of skip connections & 2 \\
LSTM activation function & Sigmoid \\
Batch size & 100 \\
Loss function & Categorical Cross entropy \\
Learning rate & $0.0001$ \\
Epochs & 100 \\
\hline
\end{tabular}
\label{tab7}
\end{table}

To assess the predictive accuracy of the hybrid CNN-LSTM model, we compare it with standalone CNN and LSTM models. All models are trained and tested using the same datasets and experimental settings. The models are implemented in Keras with a Tensorflow 2.0 backend. The training, validation, and testing data are split at a ratio of 70:10:20, respectively. The model training process involves two stages: forward propagation and backward propagation. During forward propagation, the relationship between the four input variables and the target variable (vessel ID) are established. First, the batch size is determined, followed by the initialization of the weight parameters $W$ and bias $b$. The hyperparameters are set according to Table \ref{tab:3}. The output layer of the network architecture employs the Sigmoid activation function, as shown in Equation \ref{eqn:8}.

\begin{equation}\label{eqn:8}
f(x)=\frac{1}{1+\mathrm{e}^{-x}}.
\end{equation}

Here, x implies the input value. The backpropagation approach is employed to optimize the relationship between the input and target variable. The weight parameters in the network are adjusted to minimize the discrepancy between the model prediction and the actual target value. The optimization process involves minimizing the loss, specifically the categorical cross-entropy loss function ($L_{CE}$).  This function quantifies the difference between the true distribution and the predicted distribution of class labels in a multiclass classification problem. It is defined as the negative log-likelihood of the true class, given the predicted probabilities of all classes \cite{chollet2021deep}. The mathematical expression for the categorical cross-entropy loss function is given as follows \cite{Medium}:

\begin{equation}
L_{CE} = -\sum_{i=1}^{C} y_i \cdot log(\widehat{y_i})
\end{equation}

where $y_i$ is the true label for class $i$ and $\hat{y_i}$ is the predicted probability for class $i$, and $C$ is the total number of classes.

In this study, the Adam optimizer \cite{kingma2014adam} is utilized to adjust the weights and biases during backpropagation. It is a stochastic gradient-based optimization method that adjusts the learning rate dynamically for each parameter, based on the gradient's historical information. The algorithm calculates adaptive learning rates for each parameter using the gradient's moving average and second moment.

%% The Appendices part is started with the command \appendix;
%% appendix sections are then done as normal sections

%%%%%%%%%%%%%%%%%%%%%%%%%%%%%%Discussion and results 
\section{Results and Discussion}

To evaluate the effectiveness of our model, we generate the confusion matrix, which compares the actual classes with the predicted ones. The diagonal entries of the matrix represent the true positive predictions made by the model. However, relying solely on the confusion matrix is not enough to accurately quantify the performance of the model. Therefore, we compute several commonly used evaluation metrics to fully assess the model's performance, including:

\begin{equation}
\text { Sensitivity }=\frac{\mathrm{TP}}{\mathrm{TP}+\mathrm{FN}}
\end{equation} 
\begin{equation}
\text { Specificity }=\frac{\mathrm{TN}}{\mathrm{TN}+\mathrm{FP}}
\end{equation} 
\begin{equation}
\text { F1 score }=\frac{\text { Sensitivity }+\text { Specificity }}{2} 
\end{equation} 
\begin{equation}
\text { Accuracy }=\frac{\mathrm{TP}+\mathrm{TN}}{\mathrm{TP}+\mathrm{FP}+\mathrm{FN}+\mathrm{TN}}
\end{equation}

The abbreviations TP, TN, FP, and FN correspond to true positive, true negative, false positive, and false negative predictions, respectively. Since the dataset has a class imbalance, we calculate the micro-average value for all metrics, which aggregates the contributions of all classes to determine the average metric.

\begin{figure}[!htb]
\centering
\includegraphics[width=10.5 cm]{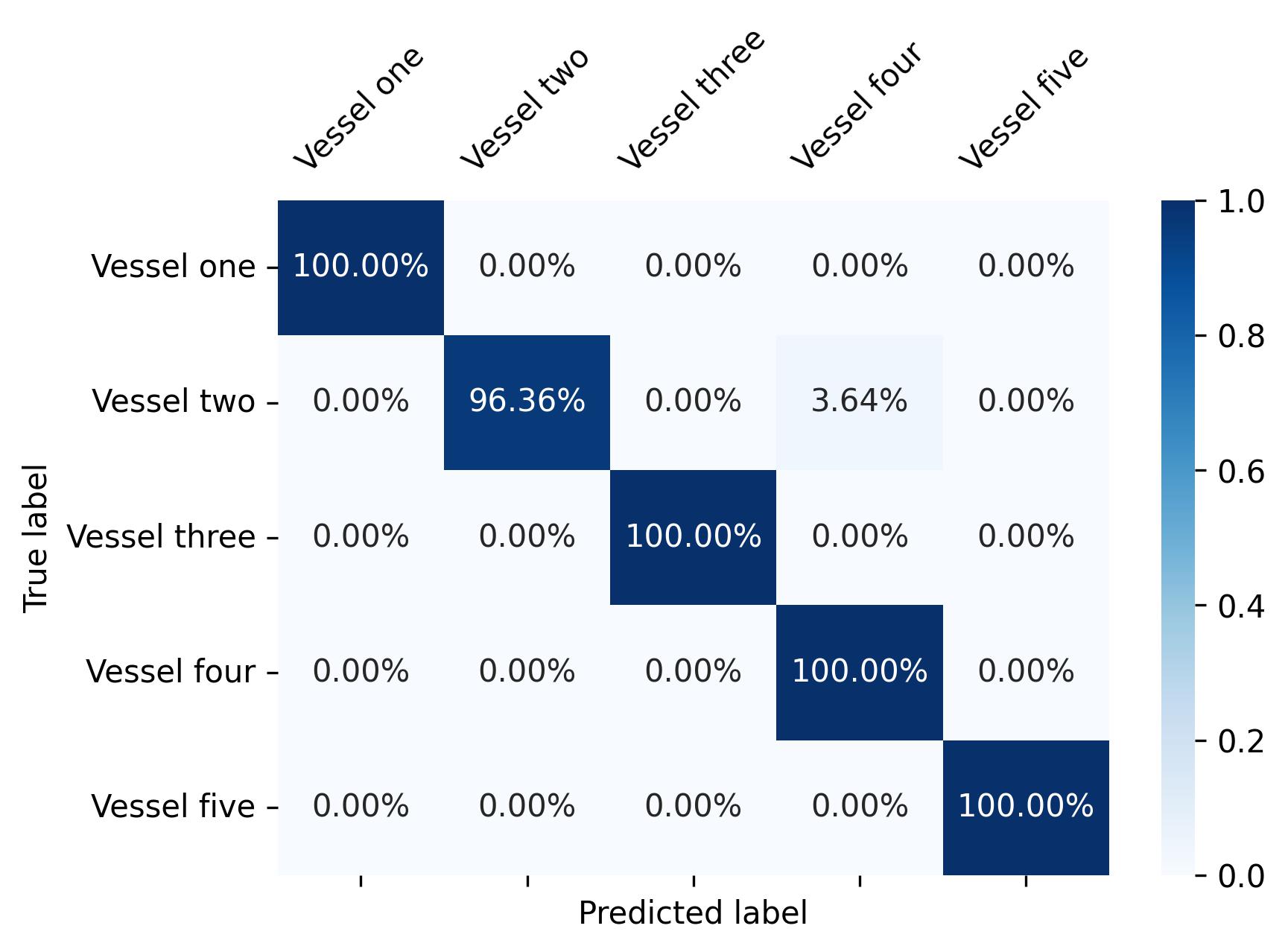}
%%\caption{Long Short Term Memory Architecture\label{fig2}}
\caption{Confusion matrix implementing CNN-LSTM architecture for small scale dataset}
\label{figure:6}
\end{figure}

To evaluate the performance of a 1D-CNN LSTM, at first, we start with a small-scale dataset consisting of five vessels. This dataset provides 5688 data points for training and validation, and 1422 data points for testing. The performance of the model is evaluated using a confusion matrix, as shown in Figure \ref{figure:6}. The results reveal remarkable accuracy for the small-scale model, with the exception of a single vessel with an accuracy of 96.4\%. The exceptional accuracy can be attributed to the diverse routes taken by each vessel, making them easily distinguishable from one another. 

In order to test the adaptability and versatility of the model, we also conduct a large-scale experiment. This comprehensive evaluation of the 1D-CNN LSTM model is designed to address two important aspects of real-world scenarios: the variation in data size and frequency of data collection among vessels, as well as the presence of overlapping tracks. 

\begin{comment}
\begin{figure}[!hb]\label{figure:11}
\centering
\includegraphics[width=13cm, height =7cm]{images/Unknowndata.png}\\
(a) Unclassified vessels \\
\vspace{.5cm}
\includegraphics[width=13cm, height =7cm]{images/predicted_data.png}\\
(b) Track association using the proposed algorithm  \\

\caption{Tracking performance of our algorithm. Thicker white dots in Figure 7(a) represent the unknown objects and Figure 7(b) depicts the assignment of the unknown vessels to their actual track using the proposed algorithm. Different color indicates unique vessels and their tracks.}
\label{figure:9}
\end{figure}
\end{comment}

\begin{figure} [!htb]
\begin{adjustwidth}{-\extralength}{0cm}
\begin{tcolorbox}[colframe=black, sharp corners]
    \centering
    \begin{subfigure}{0.49\textwidth}
        \centering
        \includegraphics[width=\linewidth]{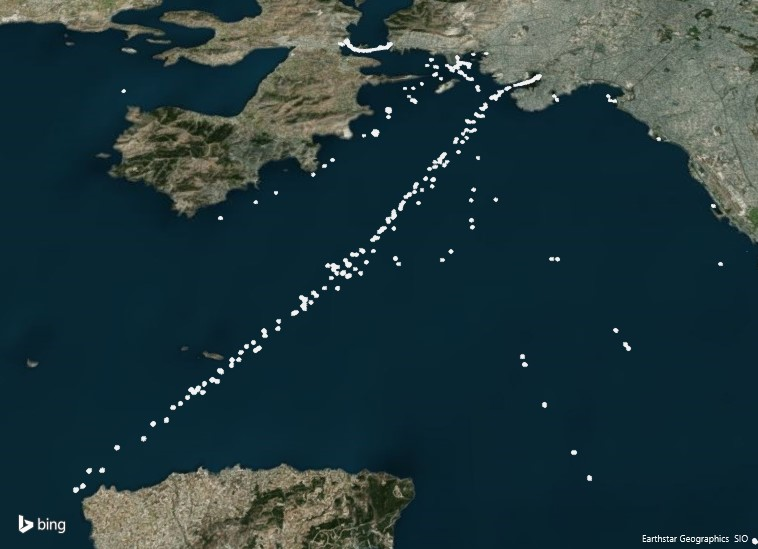}
        \caption{Unclassified vessels}
        \label{fig:subfigure1}
    \end{subfigure}%
    \hspace{5 pt}
    \begin{subfigure}{0.49\textwidth}
        \centering
        \includegraphics[width=\linewidth]{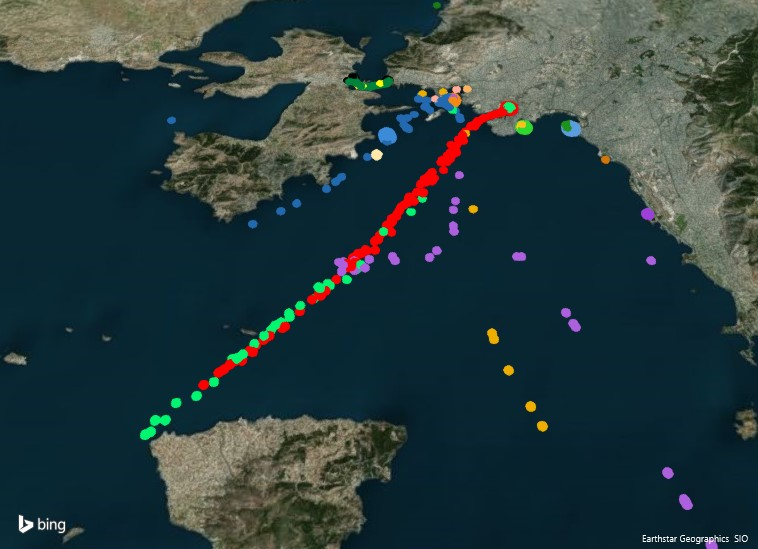}
        \caption{Track association using the proposed algorithm }
        \label{fig:subfigure2}
    \end{subfigure}
    \caption{Tracking performance of our algorithm. Thicker white dots in Figure \ref{fig:subfigure1} represent the unknown objects and \ref{fig:subfigure2} depicts the assignment of the unknown vessels to their actual track using the proposed algorithm. Different color indicates unique vessels and their tracks.}
    \label{fig:common_label}
\end{tcolorbox}
\end{adjustwidth}
\end{figure}

\begin{figure}[!ht]
\centering
\includegraphics[width=12.5 cm]{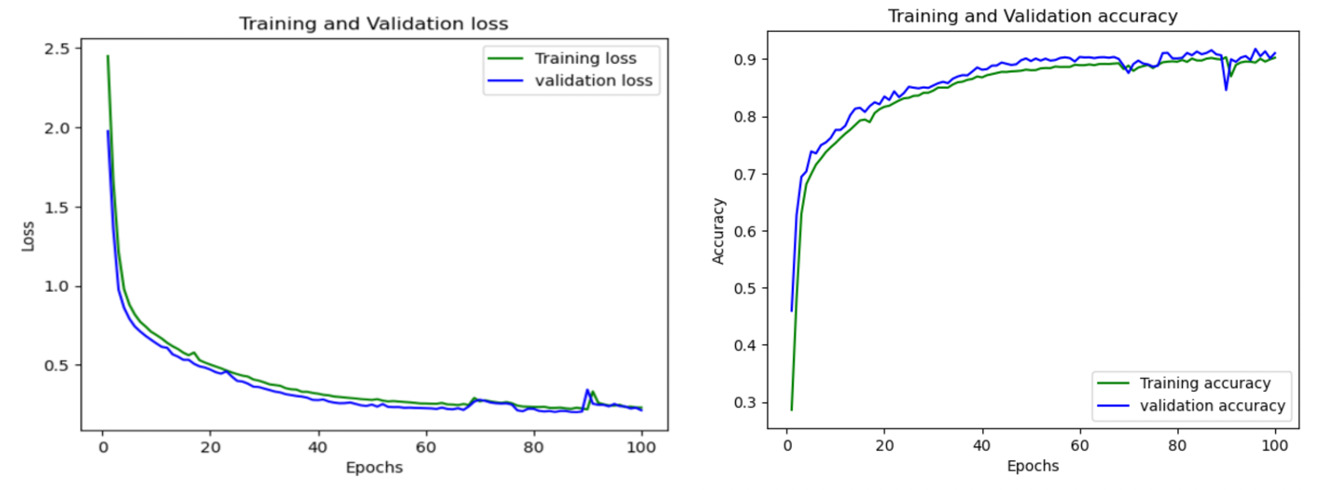}
%%\caption{Long Short Term Memory Architecture\label{fig2}}
\caption{Training and validation loss and accuracy curve for CNN-LSTM architecture}
\label{fig6}
\end{figure}
For this experiment, a total of 30 vessels are chosen, and the training and validation dataset consists of 25301 data points, while the testing dataset comprises 5060 data points. This study provides valuable insights into the adaptability of the 1D-CNN LSTM model in complex real-world scenarios and highlights its potential for practical applications. The complexity of the full-scale evaluation of the 1D-CNN LSTM model is illustrated in Figure 7(a), where vessel instances without any identifiable information are represented by white dots. There are instances of overlapping tracks with no indication of the start and end points. However, as demonstrated in Figure 7(b), our model is capable of accurately detecting these vessels and recovering their tracks, even in cases of overlapping navigation routes. The exceptional performance of 1D CNN-LSTM is also evident from the stability of its AUC curve for training and validation loss and accuracy, as shown in Figure \ref{fig6}. The model's performance can also be quantitatively assessed by comparing its predictions with the true labels.

\begin{figure}[!h]
\centering
\includegraphics[width=9.5 cm]{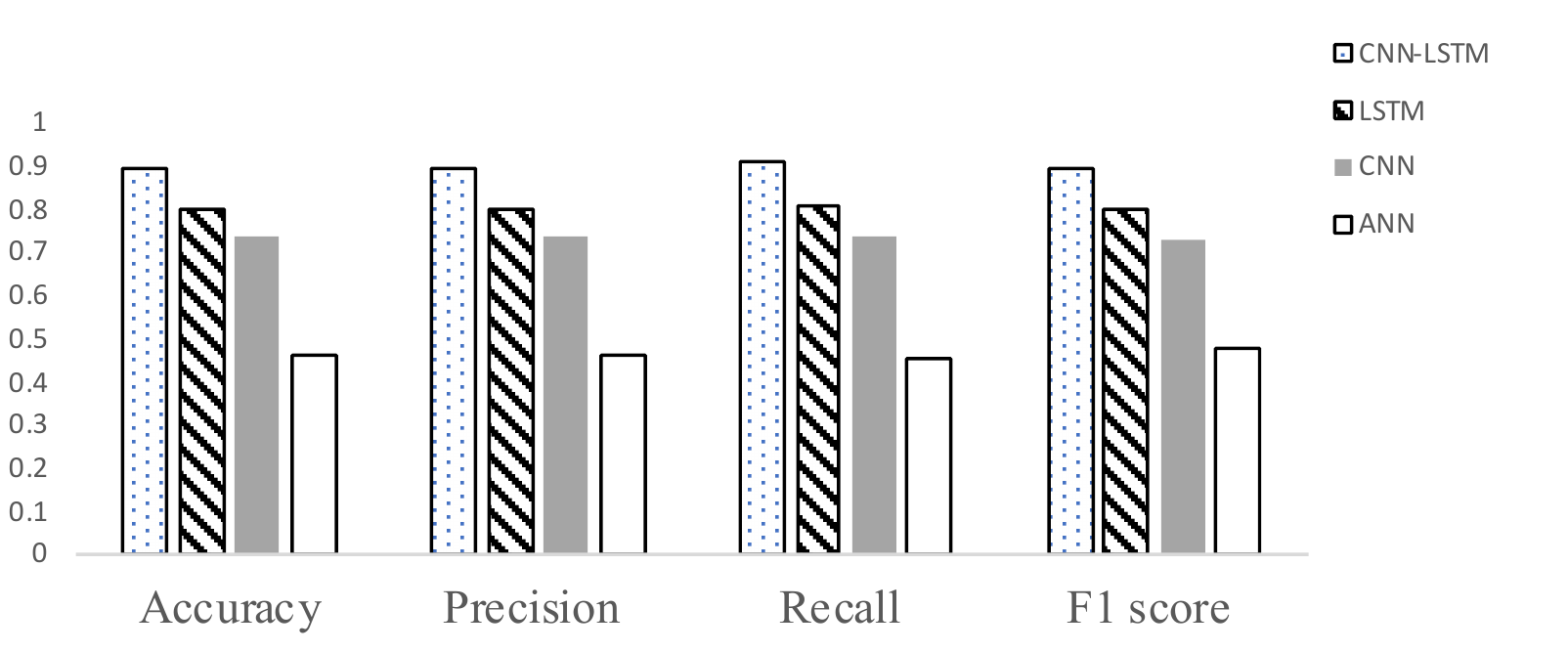}
%%\caption{Long Short Term Memory Architecture\label{fig2}}
\caption{Comparative performance of several deep learning architectures}
\label{figure:8}
\end{figure}

For the detailed quantitative evaluation using the performance metrics introduced at the beginning of this section, we apply three other competing deep learning architectures on the same AIS dataset. These approaches are regular CNN, LSTM, and feedforward artificial neural network (ANN). The 1D CNN-LSTM architecture outperformed the CNN, LSTM, and ANN architectures, achieving an accuracy score of 0.89, a precision score of 0.89, a recall score of 0.91, and an F1 score of 0.89, as shown in Table \ref{tab4}. Figure \ref{figure:8} depicts a visual summary of Table \ref{tab4}. While the LSTM model performed better than the CNN model, with an accuracy of 80\%, the ANN model exhibited the lowest accuracy, at 46\%. Given the lack of provision to account for the spatial and temporal patterns in the data, the poor performance of ANN is not surprising.

\begin{table}[ht!]
\caption {Evaluation matrix for the deep learning architectures}
\centering
    \begin{tabular}{|c|cccc|}
    \hline
        \textbf{Deep learning models } & \textbf{Accuracy} & \textbf{Precision } & \textbf{Recall } & \textbf{F1 score } \\ \hline
        CNN-LSTM & 0.89 & 0.89 & 0.91 & 0.89 \\ 
        LSTM  & 0.80 & 0.80 & 0.81 & 0.8 \\ 
        CNN & 0.74 & 0.74 & 0.74 & 0.73 \\ 
        ANN & 0.46 & 0.46 & 0.45 & 0.48 \\ \hline
    \end{tabular}
\label{tab4}
\end{table}

The superior performance of the 1D CNN-LSTM architecture is due to its ability to consider the sequential behavior of the AIS dataset, which is essential for accurate prediction in time series datasets. The architecture takes into account both the spatial and temporal characteristics of the dataset, resulting in exceptional performance in all performance evaluation metrics. Taken together, these findings underscore the efficacy of deep learning models in addressing intricate maritime problems and emphasize the potential of the 1D CNN-LSTM architecture for future tracking applications.

%%%%%%%%%%%%%%%%%%%%%%%%%%%%%%%%%%%%%%%%%%
%%%%%%%%%%%%%%%%%%%%%%%%conclusion
\section{Conclusion}
Marine vessel track association plays a critical role in national security and surveillance by enabling law enforcement agencies like the Coast Guard to detect, monitor, and track potential threat vessels. The availability of large spatiotemporal AIS datasets has provided researchers with the opportunity to develop and evaluate advanced tracking algorithms. In this study, we propose a 1D CNN-LSTM architecture for track association that leverages the strengths of two distinct neural networks to model the spatial and temporal aspects of the AIS dataset. Our numerical study demonstrates that our approach outperforms other deep learning architectures.

%test

However, like other machine learning models, our approach is unable to identify new vessels that may appear during the testing phase. One possible solution is to label these new vessels as anomalies and classify them at the end of the association process. While deep learning models have shown improved accuracy and performance, physics-based heuristics can provide a more comprehensive understanding of the underlying tracks and their dynamics. Therefore, a hybrid model that combines the strengths of both approaches could be a promising avenue for future research. 

%%%%%%%%%%%%%%%%%%%%%%%%%%%%%%%%%%%%%%%%%%

%%%%%%%%%%%%%%%%%%%%%%%%%%%%%%%%%%%%%%%%%%
\vspace{6pt} 
\reftitle{References}
\bibliography{cas-refs}

\end{document}